\documentclass[runningheads]{llncs}
\usepackage{graphicx}

\usepackage{tikz}
\usepackage{comment}
\usepackage{amsmath,amssymb} 
\usepackage{color}
\usepackage{orcidlink}

\usepackage{mathtools}
\usepackage{enumitem}
\setitemize{noitemsep,topsep=0pt,parsep=0pt,partopsep=0pt}
\usepackage{multirow}
\usepackage{booktabs}
\usepackage{color} 
\usepackage[normalem]{ulem}
\usepackage{multirow}
\usepackage{graphicx}
\usepackage{color}

\def\figa#1{Fig. \ref{fig:#1}}

\def\tab#1{Table \ref{tab:#1}}

\def\sect#1{Section \ref{sec:#1}}

\def\bestviewed{\textit{Best viewed in colour.}}

\usepackage{pifont}
\usepackage{soul}

\usepackage[accsupp]{axessibility}  

\begin{document}
\pagestyle{headings}
\mainmatter
\def\ECCVSubNumber{5559}  

\title{TDViT: Temporal Dilated Video Transformer for Dense Video Tasks}

\titlerunning{Temporal Dilated Video Transformer}
%
\author{Guanxiong Sun\inst{1,2}
\orcidlink{0000-0003-1901-9097} 
\and
Yang Hua\inst{1}
\orcidlink{0000-0001-5536-503X} 
\and
Guosheng Hu\inst{2}
\orcidlink{0000-0002-9448-9892} 
\and
Neil Robertson\inst{1}
\orcidlink{0000-0003-2461-8799}
}
\authorrunning{Guanxiong et al.}
%
\institute{EEECS/ECIT, Queen's University Belfast, UK \and
Oosto, Belfast, UK \\
\email{\{gsun02, y.hua, n.robertson\}@qub.ac.uk}, 
\email{huguosheng100@gmail.com}
}
\maketitle

\begin{abstract}

Deep video models, for example, 3D CNNs or video transformers, have achieved promising performance on sparse video tasks, i.e., predicting one result per video. However, challenges arise when adapting existing deep video models to dense video tasks, i.e., predicting one result per frame. Specifically, these models are expensive for deployment, less effective when handling redundant frames and difficult to capture long-range temporal correlations. To overcome these issues, we propose a Temporal Dilated Video Transformer (TDViT) that consists of carefully-designed temporal dilated transformer blocks (TDTB). TDTB can efficiently extract spatiotemporal representations and effectively alleviate the negative effect of temporal redundancy. Furthermore, by using hierarchical TDTBs, our approach obtains an exponentially expanded temporal receptive field and therefore can model long-range dynamics. Extensive experiments are conducted on two different dense video benchmarks, i.e., ImageNet VID for video object detection and YouTube VIS for video instance segmentation. Excellent experimental results demonstrate the superior efficiency, effectiveness, and compatibility of our method.  The code is available at \href{https://github.com/guanxiongsun/vfe.pytorch}{https://github.com/guanxiongsun/vfe.pytorch}.

\end{abstract}

\section{Introduction}
\label{sec:intro}

In the past decade, 2D convolutional neural network (CNN) based architectures \cite{resnet,googlenetv4,inception,googlenetv3,effnet,resnext} have dominated various computer vision tasks for still image understanding, e.g., image classification, object detection and semantic segmentation.
Given the excellent performance on still images, 2D CNNs are adapted to video understanding by extending the 2D convolutional layer through the temporal axis, i.e., 3D CNNs  \cite{i3d,p3d,3dcnns}.
Recently, the computer vision community has seen a model shift from conventional CNNs to vision transformers  \cite{vit,attention}. This trend began with the pioneering approach, ViT  \cite{vit}, which leverages Transformers  \cite{attention} to understand an image as a sequence of non-overlapping patches. Some follow-up approaches focus on improving ViT in different aspects, such as finer feature representations \cite{tit,swin,pvt}, better token generations \cite{cvt,t2t}, more efficient training processes \cite{deit}, etc. Not surprisingly, the great success of transformers on still-image tasks also leads to 
investigation of the use of transformers for video tasks \cite{vivit,timesformer,videoswin}.

\begin{figure}[t]
  \centering
  \includegraphics[width = 0.9\columnwidth]{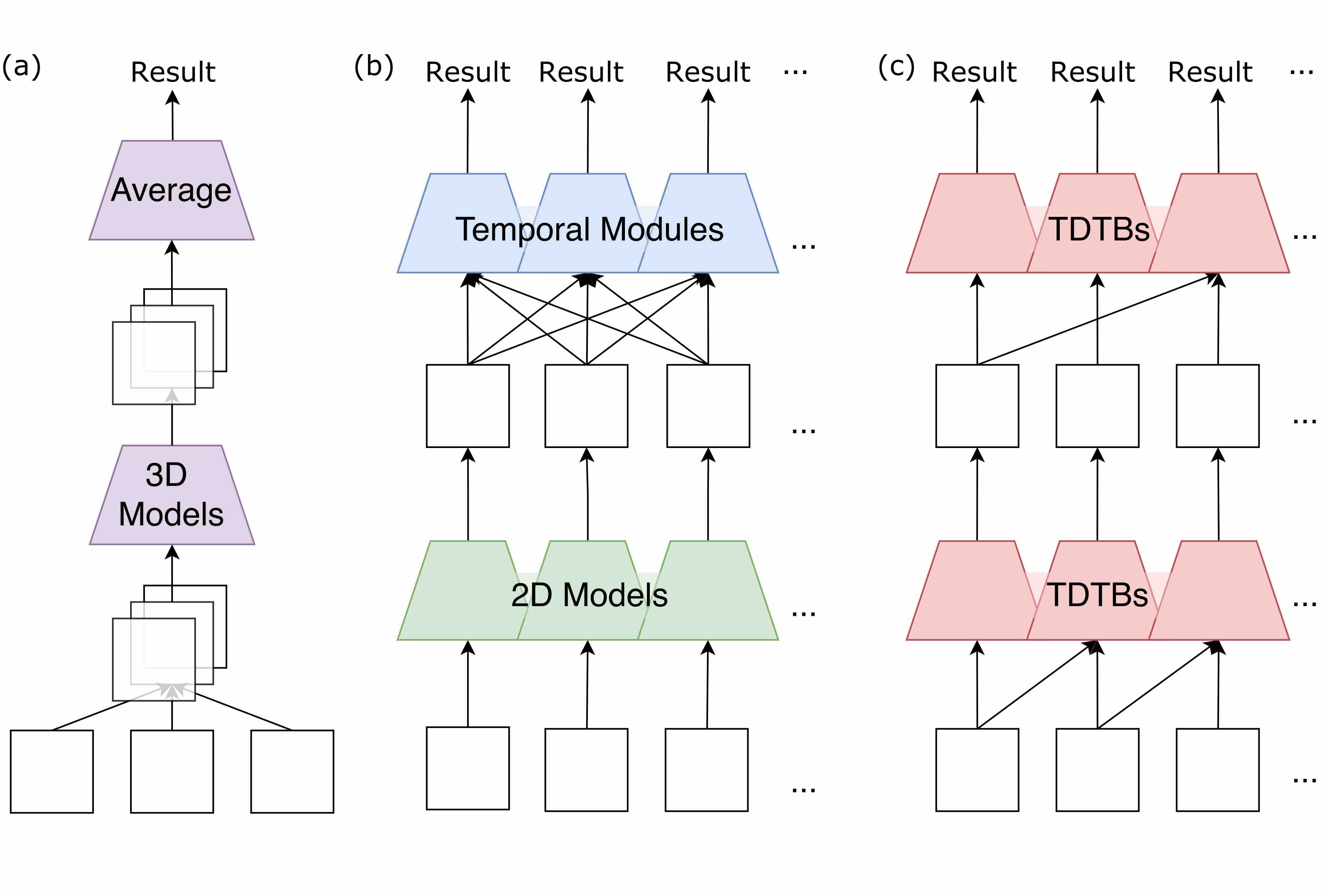}
\caption{
Architecture (a) is widely used in sparse video tasks. 3D models, e.g., 3D CNNs, take multiple frames as input and generate one output by averaging the spatiotemporal representations. Architecture (b) is used for dense video tasks. Considering the computational cost, 2D models are used to extract features of independent frames, and then temporal modules, e.g., correlation filters, are leveraged to model spatiotemporal correspondences. Our TDViT (c) is designed for dense video tasks, which can efficiently and effectively extract spatiotemporal representations using temporal dilated transformer blocks (TDTB). \bestviewed
}
  \label{fig:intro}
\end{figure}

Most video models (3D CNNs \cite{i3d,slowfast,3dcnns} or video transformers  \cite{timesformer,omni,videoswin}) are designed for \emph{sparse} video tasks, e.g., video classification \cite{youtube8m,videoclassification,vccscnn} and action recognition  \cite{i3d,activitynet,s2s}, whose goal is to predict one label for one given video. These models utilise 3D operations, e.g., 3D convolutional layers or spatiotemporal self-attentions, to learn a spatiotemporal representation for a video. The architecture of sparse video models is shown in \figa{intro}(a). 
Differently, \emph{dense} video tasks aim to predict one result per frame, e.g., video object detection \cite{mega,mamba} and video instance segmentation \cite{sipmask,youtubevis}.
However, problems arise when integrating video models into dense video tasks. Specifically, these models are (1) computationally intensive, resulting in high deployment costs; (2) unable to extract relevant information from redundant frames; and (3) challenging to capture long-range temporal correlations.
In practise, therefore, 
state-of-the-art approaches for dense video tasks are still built in a hybrid manner: extracting spatial features using 2D models and adding temporal modules \cite{flownet,attention} to generate spatiotemporal representations. The architecture of sparse models is shown in \figa{intro}(b).

In this paper, we propose a Temporal Dilated Video Transformer (TDViT) whose overall architecture is shown in \figa{intro}(c).
Inspired by visual transformers \cite{vit,swin,deit} which are naturally suitable for sequence modelling, we exploit transformers to extract spatiotemporal features. 
Unlike most video models that are based on self-attention
transformers, we present a novel temporal dilated transformer block (TDTB) with two distinct designs. Firstly, a memory structure that stores features of previous frames is introduced to each TDTB. 
During inference, the query tokens are derived from the current frame for each time step, whereas the key and value tokens are sampled from the memory structure.
In this way, TDTB can extract multi-frame spatiotemporal features with a single-frame overhead. 
Secondly, a temporal dilation factor is proposed to control the temporal frequency of feature sampling by skipping neighbouring frames. Therefore, TDTB can alleviate the negative effects of video redundancy.
Furthermore, by employing hierarchical TDTBs in stages, 
TDViT acquires an exponentially expanded temporal receptive field. Here, the temporal receptive field represents the number of video frames considered when extracting spatiotemporal features. 
As a result, TDViT can effectively capture long-range temporal correlations and achieve higher accuracy.

To better demonstrate the differences between our TDViT and other visual models, we categorise existing visual models based on their architectures and their target tasks. For example, 
ResNets  \cite{resnet,resnext} and Inceptions  \cite{inception,googlenetv3} are CNN-based models designed for image tasks. VSwin  \cite{videoswin} and TSFMR  \cite{timesformer} are transformer-based models designed for sparse video tasks. Our approach is also a transformer based model but designed for dense video tasks. The details and categorisation of existing video models are presented in \tab{category}.

To sum up, our contributions are listed as follows:
\begin{itemize}
  \item
  Unlike the existing hybrid solutions (2D image models equipped with temporal modules \cite{flownet,attention}) for dense video tasks, we propose a neat and compact architecture, 
  Temporal Dilated Video Transformers (TDViT), for spatiotemporal representation learning.

  \item We design a novel temporal dilated transformer block (TDTB) that can efficiently extract spatiotemporal representations and alleviate the negative effect of video redundancy by cooperating with a memory structure and temporal dilated feature sampling process. Furthermore, by employing hierarchical TDTBs, our approach can effectively capture long-range temporal correlations.

  \item We conduct extensive experiments on two datasets for different dense video tasks, i.e., ImageNet VID for video object detection and YouTube VIS for video instance segmentation. Excellent experimental results demonstrate the superior effectiveness, efficiency, and compatibility of our method. Given these advantages, we believe that our TDViT can serve as a general-purpose backbone for various dense video tasks. 
\end{itemize}

\begin{table}[t]
\caption{Categorization of visual models based on their architectures and their target tasks. For dense video tasks, most SOTA methods use 2D models as their backbone and leverage temporal modules, e.g., attention \cite{attention} or correlation filter \cite{flownet} to get spatiotemporal representations. To the best of our knowledge, our approach is the first end-to-end backbone designed for dense video tasks.}
\label{tab:category}
\centering
\begin{tabular}{cccc}
\toprule
Arch. & 2D Models & 3D Sparse Models & 3D Dense Models \\
                       \midrule
CNN &
  \begin{tabular}[c]{@{}c@{}}ResNet  \cite{resnet,resnext}\\ Inception  \cite{inception,googlenetv3}\\ EffNet  \cite{effnet}\end{tabular} &
  \begin{tabular}[c]{@{}c@{}}R3D  \cite{r3d}\\ I3D  \cite{i3d}\\ SFNet  \cite{slowfast}\end{tabular} &
  \begin{tabular}[c]{@{}c@{}}2D models +\\ Temporal modules \cite{flownet,attention} \end{tabular} \\
  \midrule
Transformer &
  \begin{tabular}[c]{@{}c@{}}ViT  \cite{vit} \\ DeiT  \cite{deit}\\ Swin  \cite{swin}\end{tabular} &
  \begin{tabular}[c]{@{}c@{}}Non-local  \cite{nonlocal}\\ TimeSformer \cite{timesformer}\\ VSwin  \cite{videoswin}\end{tabular} &
  \begin{tabular}[c]{@{}c@{}}TDViT\\ (Ours)\end{tabular}\\
  \bottomrule
\end{tabular}
\end{table}

\begin{figure}[t]
  \centering
  \includegraphics[width = \textwidth]{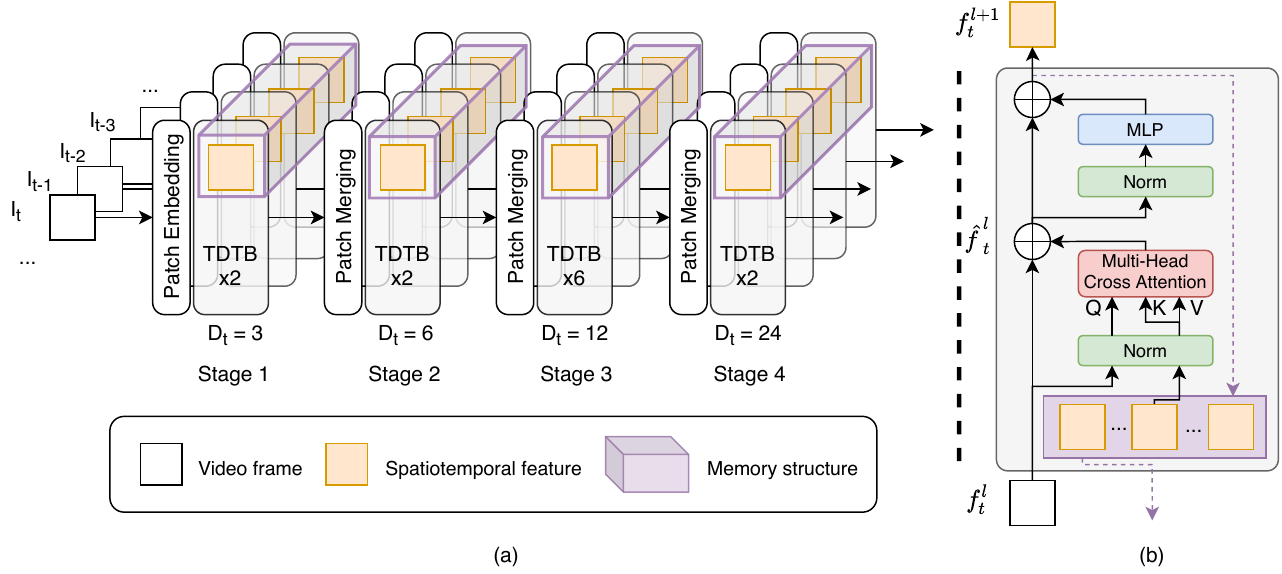}
\caption{
(a) Overview. TDViT contains four stages which consist of several temporal dilated transformer blocks (TDTB). A memory structure (purple cuboids) is introduced into the TDTB, which stores features of previous frames (yellow rectangles) and enables our approach to dynamically establish temporal connections. The temporal dilation factor $D_t$ is used to control the memory sampling process and reduce the video redundancy. (b) Details of a TDTB. For every time step, the query tokens are from the current frame $I_t$ but the key and value tokens are derived from memory sampling. Finally, the memory structure saves the output features and deletes the oldest features. \bestviewed
}
  \label{fig:overall}
\end{figure}

\section{Related Work}
\label{sec:related}

\subsubsection{Video Models for Sparse Video Tasks.} 
For video modelling, 3D CNNs are investigated in pioneering approaches \cite{i3d,p3d,3dcnns}.
These methods extend 2D CNNs through the temporal axis, but the performance is limited by their small temporal receptive fields.
To address this, multi-head attentions  \cite{attention} are introduced to fuse spatiotemporal representations. For example, NLNet  \cite{nonlocal} adopts self-attentions to model pixel-level dependency globally. GCNet  \cite{gcnet} presents a lightweight global context block and achieves good performance with less computation. DNL  \cite{dnlnet} captures better visual clues using disentangled NL blocks.

\subsubsection{Dense Video Tasks.} Dense video tasks aim to output one prediction per frame.
For video object detection, DFF  \cite{dff}, FGFA  \cite{fgfa} and THP  \cite{thp} use optical flow to propagate or aggregate feature maps of neighbour frames. Recent methods, SELSA \cite{selsa}, RDN  \cite{rdn} and MEGA \cite{mega}, perform proposal-level aggregation by modelling relationships between objects using attention mechanisms \cite{attention}.
For video instance segmentation, MaskTrack R-CNN  \cite{youtubevis} extends instance segmentation from image domain to video domain. MaskProp  \cite{maskprop} uses mask propagation to obtain smoother results. StemSeg  \cite{stem-seg} treats the video clip as 3D spatial-temporal volume and segments instances by clustering methods.
TransVOD  \cite{transvod} and VisTR  \cite{vistr} adapt DETR  \cite{detr} for VOD and VIS tasks, respectively, in a query-based end-to-end fashion.
These methods are built in a hybrid manner, extracting features with 2D models and relying on additional temporal modules. In contrast, our TDViT is a neat and compact model that is built with unified TDTBs.

\subsubsection{Vision Transformers.}
Our approach is inspired by the recent trend of attention based vision transformers. Among these methods, 
ViT  \cite{vit} is the first  to use consecutive transformers to replace the CNN backbones.
It reshapes an image into a sequence of non-overlapping flattened 2D patches and achieves a good speed-accuracy trade-off for image classification.
Many recent papers improve the performance of ViT in different aspects. For example, Deit \cite{deit} addresses the issue that ViT requires a large amount of data by knowledge distillation. TiT  \cite{tit} and PVT \cite{pvt} generate multi-scale feature maps to capture finer visual clues. T2T  \cite{t2t} uses a small transformer network to generate better token representations.
The most related work to ours is the Swin  \cite{swin}, which uses local attention to gradually reduce the resolution of feature maps and achieves impressive performance in various vision tasks.
These methods demonstrate that transformers do well in modelling spatial dependencies.
Inspired by these methods, we dive deeper into the essence of transformers and find that they are naturally suitable for modelling the correlations of sequence data.



\section{Method}
\label{sec:method}

\subsection{Overall Architecture}
An overview of our TDViT is presented in \figa{overall}(a).
Our approach consists of four stages. Each stage is obtained by stacking several temporal dilated transformer blocks (TDTB). 
Each TDTB has a memory structure to store and sample features of previous frames.
TDViT takes a frame $I_t$ of one video $\{I_\tau\}_{\tau=1}^T$ as input, where $T$ is the length of the video and $t$ denotes the current time step. The previous frames of the video can be denoted as $\{I_\tau\}_{\tau=1}^{t-1}$. 
By modelling the spatiotemporal correlations between $I_t$ and $\{I_\tau\}_{\tau=1}^{t-1}$, TDViT can output more accurate predictions for every frame.

\subsection{Temporal Dilated Transformer Block}
\label{subsect:tdtb}
Inspired by the dilated convolutions which introduce spatial skip steps for feature aggregation, we propose the temporal dilated transformer block to construct skip connections in the temporal dimension.
Specifically, unlike existing self-attention based transformer modules, the temporal dilated transformer block (TDTB) has two novel designs: (1) a memory structure $\mathcal{M}$ to explicitly store previous features; (2) a temporal dilation factor $D_t$ to control the stride of temporal connections.
The input of a TDTB is the features of the current frame, which can be in different forms, such as feature maps with arbitrary sizes or proposal features.
Given $D_t$=$d$, the $l$-th TDTB samples features from $\{I\tau\}_{t-d}^t$ as reference features $f^R$. Then $I_t$ is enhanced with $f^R$ using multi-head cross attention (MCA). Specifically, the tokens for query $Q$, key $K$ and value $V$ are obtained by normalising features of the current frame $f^l_t$ and reference features $f^R$, respectively. A 2-layer MLP with GELU non-linearity is applied to the MCA's output features $\hat{f}^{l}_t$. Two LayerNorm $LN$ layers are applied before MCA and MLP, respectively. The output feature $f^{l+1}_t$ is obtained by applying residual connections after MCA and MLP layers. Formally,

\begin{equation}
\begin{aligned}
& f^R = Sampling(\mathcal{M}, D_t) \text{,}\\
& \hat{f}^{l}_t = MCA(LN(f^{l}_t), LN(f^R)) + f^{l}_t \text{,}\\
& f^{l+1}_t = MLP(LN(\hat{f}^{l}_t)) + \hat{f}^{l}_t \text{,}
\end{aligned}
\end{equation}
where $Sampling(\cdot, \cdot)$ denotes the sampling process guided by $D_t$. An illustration of TDTB is shown in \figa{overall} (b).

\begin{figure}[t]
  \centering
  \includegraphics[width = 0.8\columnwidth]{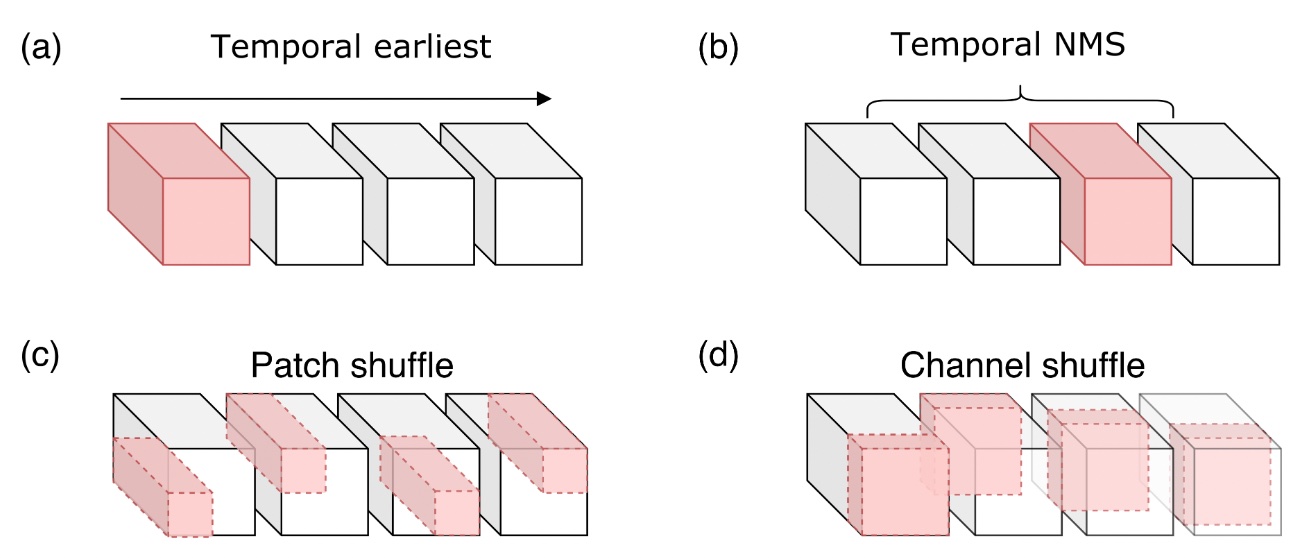}
\caption{
Illustration of memory sampling. The red cuboid denotes the sampled features and the grey cuboid denotes stored features in memory. \bestviewed
}
  \label{fig:sampling}
\end{figure}

\subsubsection{Memory Sampling and Feature Reuse.} 

The memory structure $\mathcal{M}$ is updated at every time step by replacing the oldest frame with $f^{l+1}_t$. Given a temporal dilation $D_t$, we design different strategies to sample reference features $f^R$.
The temporal \textit{earliest} strategy selects the oldest feature map in the memory. The temporal NMS selects the feature map that has the maximum L2 norm. 
The $patch$-wise shuffle strategy first splits the feature maps into four windows, i.e., top left, top right, bottom left, and bottom right. Then, it randomly selects four time steps from four groups, respectively, and combines them as the reference feature map.
Similarly, the $channel$-wise shuffle strategy split the feature maps into $C$ groups according to channels, where $C$ is the number of channels. Among these strategies, the first two are feature-level operations, while the last two are more fine-grained group-wise operations.
The differences between these strategies are shown in \figa{sampling}. 

Once we obtain the sample features $f^R$, we  reuse the feature for $D_t$ time steps and then sample new $f^R$ from the updated memory. Instead of computing query, key, and value tokens for every frame, TDTB only extracts query tokens from the current frame and reuses $f^R$ to obtain key and value tokens. As a result, TDViT achieves a slightly faster speed compared to its 2D counterpart, the Swin \cite{swin}.

\begin{figure}[t]
  \centering
  \includegraphics[width = 0.8\columnwidth]{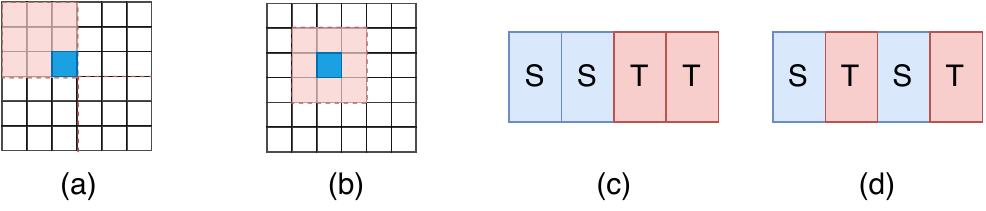}
\caption{
(a) Window and (b) correlation based local attentions. The blue rectangle denotes one query token, the red dotted rectangle denotes the range of participated key tokens. (c) Split and (d) factorised spatiotemporal schemes. The blue and red boxes denote space-only self-attention and TDTB, respectively. \bestviewed}
\label{fig:st_scheme_and_local_att}
\end{figure}

\subsubsection{Efficient Local Attentions.}
The most well-known transformers are built with multi-head self-attention (MSA) modules in which one query token attends to all key tokens globally. 
We exploit local attentions because they have lower computational costs.
In addition, local attentions are very suitable for dense video tasks since
the movement of an object is continuous and should be inside a local region for a short time. 
Specifically, we introduce two different local attentions. Firstly, following the protocol of  \cite{swin}, we adapt the window-based local attention (\figa{st_scheme_and_local_att}(a)), where key tokens are from reference frames within the same window of the query token.
Secondly, inspired by \cite{flownet,psla,flownet2}, we design a local attention based on correlation (\figa{st_scheme_and_local_att}(b)), where key tokens are from reference frames within a neighbour region of the query token.

\subsection{Spatiotemporal Attention Schemes}
\label{subsect:st_schemes}
We investigate different spatiotemporal attention schemes, i.e., the order of the space block and the temporal block (TDTB) in a stage. Inspired by \cite{timesformer,videoswin}, we design two spatiotemporal schemes: split and factorised schemes.
The split scheme adds several temporal blocks (TDTB) after consecutive space blocks (\figa{st_scheme_and_local_att}(c)), while the factorised scheme adds one TDTB after each space block as shown in \figa{st_scheme_and_local_att}(d).

\subsection{Temporal Receptive Field of Hierarchical TDTB}
\begin{figure}[t]
  \centering
  \includegraphics[width = 0.8\columnwidth]{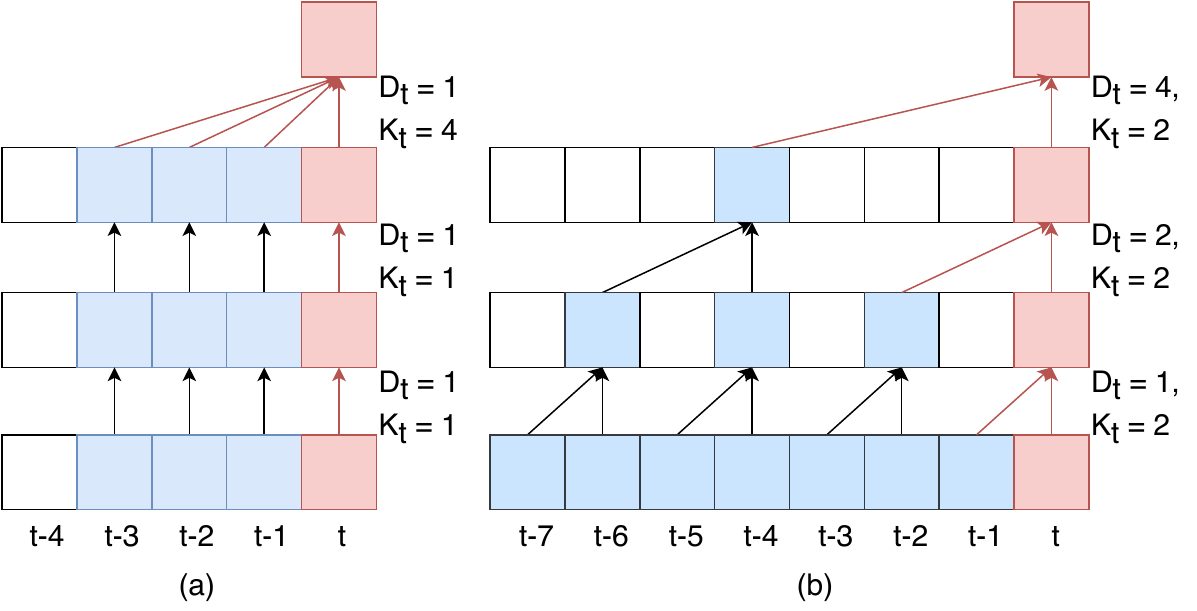}
\caption{
Illustration of how hierarchical TDTBs increase the temporal receptive field. 
(a) and (b) show the framework of RDN \cite{rdn} and our TDViT, respectively. The red and blue rectangles denote the current frame and frames within the temporal receptive field, respectively. \bestviewed
}
\label{fig:hierarchical}
\end{figure}

One TDTB with temporal dilation $D_t$ can aggregate temporal information from time step $t-{D_t}$ to $t$. We use TDTBs in multiple hierarchical stages to obtain an exponentially expanded temporal receptive field.
For a better demonstration, we introduce the temporal kernel size $K_t$ which denotes the number of frames considered in an aggregation process. For example, an aggregation process with $K_t$=2 and $D_t$=4 aggregates 2 frames whose temporal distance is 4. Using $K_t$ and $D_t$, we can describe the process of TDViT and other SOTA methods for VOD from a unified perspective.  
\figa{hierarchical}(a) shows a simplified version of RDN \cite{rdn} where $D_t$=$\{1,1,1\}$ and $K_t\text{=}\{1,1,4\}$ from bottom to top. For RDN, temporal aggregation only happens in the detection head after the backbone feature extraction.
TDViT hierarchically fuses temporal features onto the current frame. \figa{hierarchical}(b) shows a simplified version of TDViT, where $D_t$=$\{1,2,4\}$ and $K_t$=$\{2, 2,2\}$.
With similar complexity, i.e., the times of feature aggregation shown as the number of red connections in \figa{hierarchical}, the temporal receptive fields of RDN and TDViT are 4 and 8, respectively.

\subsection{Architecture and Variants}

Following the design principles of Swin \cite{swin}, we build three basic variants of our models, called TDViT-T (tiny), TDViT-S (small) and TDViT-B (base), whose sizes and computational complexities are similar to Swin-T, Swin-S and Swin-B.
The query dimension of each transformer head is 32, and the expansion layer of each MLP is 4. 
Two hyperparameters are introduced to describe each variant: $C$ denotes dimensions of the linear embedding in the first stage and $L$ denotes the number of transformers in each stage. The  \textit{basic} variants are represented as TDViT-T: $C$=96, $L$=(2,2,6,2); TDViT-S: $C$=96, $L$=(2,2,18,2); TDViT-B: $C$=128, $L$=(2,2,18,2).
By adding two additional TDTBs at the end of stage 3, we further obtain the \textit{advanced} variants which can be represented as TDViT-T$^+$: $C$=96, $L$=(2,2,8,2); TDViT-S$^+$: $C$=96, $L$=(2,2,20,2); TDViT-B$^+$: $C$=128, $L$=(2,2,20,2).



\section{Experiments}
\label{sec:exp}
To verify the generality of our TDViT, we conduct experiments on two datasets, i.e., ImageNet VID \cite{imagenet} for video object detection (VOD) and YouTube VIS \cite{youtubevis} for video instance segmentation (VIS). Firstly, we compare TDViT with other backbones. Then, we adapt TDViT to SOTA VOD and VIS methods to evaluate its compatibility. Finally, we conduct ablation studies on different designs, as mentioned in the \sect{method}. We report detailed evaluation results in the COCO \cite{coco} format. Specifically, we use the average precision (AP$^{box}$) metric which computes the average precision over ten IoU thresholds [0.5: 0.05: 0.95] for all categories. Meanwhile, we report other important metrics, e.g., AP$^{box}_{50}$ and AP$^{box}_{75}$ are calculated at IoU thresholds of 0.50 and 0.75 \cite{voc}, and AP$^{box}_{S}$, AP$^{box}_{M}$,  AP$^{box}_{L}$ are calculated on different object scales, i.e., small, medium and large. All training and testing are performed on Tesla V100 GPUs.

\subsubsection{Video Object Detection Setup.}
The ImageNetVID dataset \cite{imagenet} contains 3,862 training and 555 validation videos. Following previous approaches  \cite{mamba,selsa,fgfa}, we add overlapped 30 classes of the ImageNet DET dataset into the training set. Specifically, we sample 15 frames from each video in the VID training set and at most 2,000 images per class from the DET training set.
We employ an AdamW  \cite{adam} optimizer for 3 epochs in total, with an initial learning rate of $10^{-3}$ and $10^{-4}$ after the 2${nd}$ epoch, using a weight decay of 0.05. We follow the augmentation and regularisation strategies of Swin \cite{swin}. Given the key frame $I_k$, we randomly sample four frames from $\{I_{\tau}\}_{k-D_t}^{k+D_t}$ to approximately form the memory of TDTBs in four stages, respectively. During training, detection losses are only computed for the key frame $I_k$.

\subsubsection{Video Instance Segmentation Setup.} 
We also evaluate TDViT for video instance segmentation on YouTube VIS 2019 dataset \cite{youtubevis} and report the validation results as \cite{sipmask,stem-seg}. There are 3,471 training videos and 507 validation videos. For training and testing, the original images are resized to 640$\times$360. 
The metrics are computed on masks, denoted as AP$^{mask}$ and AP$^{mask}_{75}$, etc.

\subsection{Video Object Detection}
We first compare TDViT with other backbones using single-frame FasterRCNN \cite{fasterrcnn} as the base detector. Then, we integrate TDViT with the SELSA \cite{selsa} video object detection (VOD) method and compare it with other SOTA VOD methods.
\begin{table*}[t]
\centering
\caption{Comparisons with other widely used backbones for video object detection. }
\label{tab:backbones}
\begin{tabular}{lccccccccc} 
\toprule
Backbone    & Arch.                 & AP$^{box}$ & AP$^{box}_{50}$ & AP$^{box}_{75}$ & AP$^{box}_S$ & AP$^{box}_M$ & AP$^{box}_L$ & \#Param & FPS   \\
\midrule
R-50  \cite{resnet}        & \multirow{2}{*}{CNN}  & 44.3     & 72.5        & 47.2        & 6.3        & 19.7       & 49.3       & 44.4M  & 23.0  \\
R-101  \cite{resnet}        &                       & 48.5     & 75.5        & 53.1        & 7.6        & 23.1       & 53.7       & 63.4M  & 19.5  \\
\midrule
Swin-T  \cite{swin}     & \multirow{3}{*}{ViT} & 47.1     & 77.2        & 51.5        & 9.5        & 23.2       & 52.4       & 47.8M  & 22.8  \\
Swin-S  \cite{swin}     &                       & 52.6     & 82.4        & 59.3        & 10.0       & 26.5       & 58.6       & 69.1M  & 16.7  \\
Swin-B  \cite{swin}     &                       & 53.2     & 82.7        & 60.4        & 9.1        & 27.7       & 59.0       & 107.1M & 12.8  \\
\midrule
TDViT-T     & \multirow{6}{*}{ViT} & 49.1     & 78.5        & 52.7        & 8.0        & 25.6       & 53.1       & 47.8M  & 23.9  \\
TDViT-T$^+$ &                       & 50.9     & 79.9        & 55.7        & 9.1        & 26.9       & 57.2       & 51.3M  & 21.9  \\
TDViT-S     &                       & 55.4     & 84.1        & 63.4        & 10.3       & 29.3       & 61.2       & 69.1M  & 16.8  \\
TDViT-S$^+$ &                       & 55.7     & 84.1        & 63.2        & 10.5       & 30.4       & 61.4       & 72.7M  & 16.2  \\
TDViT-B     &                       & 56.0     & 84.4        & 64.1        & 10.7       & 28.0       & 62.3       & 107.1M & 12.9  \\
TDViT-B$^+$ &                       & 56.1     & 84.7        & 64.2        & 10.0         & 29.9       & 61.8       & 113.5M & 12.2  \\
\bottomrule
\end{tabular}
\end{table*}

\subsubsection{Comparisons with other backbones.}  We compare TDViT with various widely used backbones, e.g., R-50 and R-101 as baselines. We also introduce a transformer-based backbone, Swin \cite{swin} as the baseline for fair comparisons.
Detailed evaluation results are tested with the single-frame FasterRCNN  \cite{fasterrcnn} as shown in \tab{backbones}. All TDViT variants greatly surpass other architectures with comparable model sizes, i.e., number of parameters denoted as \#Param. For \textit{basic} variants, TDViT-T achieves 49.1\% of AP$^{box}$, outperforming R-50/Swin-T (44.3/47.1\%) by +4.8/2.0\% of AP$^{box}$, respectively. TDViT-S/B achieves 55.4/ 56.0\% of AP$^{box}$, outperforming Swin-S/B (52.6/53.2\%) by +2.8\% and TDViT-S outperforms R-101 (48.5\%) by +6.9\% of AP$^{box}$. 
Using the \textit{advanced} variants, i.e., TDViT-T$^+$/S$^+$/B$^+$, the performance is improved to 50.9/55.7/56.1\% of AP$^{box}_{50}$, respectively.
To compare the run-time speed, we test all models on the same device, and the detailed results (FPS) are listed in the last column of \tab{backbones}. 
TDViT achieves better speed-accuracy trade-offs, and the speed-accuracy curve is shown in Fig. S1 in the supplementary material.

\begin{table}[t]
\centering
\caption{Comparisons with SOTA video object detection methods.}
\label{tab:sota}
\begin{tabular}{lccccccc} 
\toprule
Methods    & Backbone & AP$^{box}_{50}$ & Slow          & Medium & Fast & FPS           & Hardware    \\ 
\midrule
FGFA     \cite{fgfa}   & R-101     & 76.3        & 83.5          & 75.8   & 57.6 & 1.2           & K40         \\
PSLA  \cite{psla}       & R-101+DCN & 80.0        & -             & -      & -    & 26.0          & Titan V     \\
CHP  \cite{chp}       & R-101     & 76.7        & -             & -      & -    & 37.0 & -           \\
TransVOD  \cite{transvod}   & R-101     & 81.9        & -             & -      & -    & -             & -           \\
SELSA   \cite{selsa}    & R-101     & 80.3        & 86.9          & 78.9   & 61.4 & -             & -           \\
RDN  \cite{rdn}       & R-101     & 81.8        & 89.0          & 80.0   & 59.5 & 10.6          & Tesla V100  \\
MEGA  \cite{mega}       & R-101     & 82.9        & 89.4 & 81.6   & 62.7 & 8.7           & 2080 Ti     \\
TFBlender  \cite{tfblender} & R-101     & 83.8        & -             & -      & -    & 4.9           & 2080 Ti     \\ 
MAMBA \cite{mamba} & R-101     & 84.6        & -             & -      & -    & 9.1          & Titan RTX   \\ 
\midrule
SELSA*  & R-101  &81.5   &87.8   & 79.9  & 65.4  & 14.5 &Tesla V100 \\
SELSA+Ours & TDViT-T  & 83.9        & 88.6          & 83.8   & 67.7 & 16.2          & Tesla V100  \\
SELSA+Ours & TDViT-T$^+$ & 84.5        & 88.9          & 84.3   & 69.4 & 15.7          & Tesla V100  \\
\bottomrule
\end{tabular}
\end{table}

\subsubsection{Comparisons with SOTA.} 
SOTA VOD methods use multi-frame proposal aggregation to learn the relationships of objects in a video and achieve promising performance. 
We use SELSA \cite{selsa} as our baseline for simplicity. In our re-implementation of SELSA, we introduce the distillation strategy proposed in RDN \cite{rdn} where the top 75 out of total 300 proposals on the reference frames are selected for conducting aggregation. 
We follow previous methods  \cite{mega,rdn,mamba,fgfa} and use Average Precision with IoU threshold 0.5 (AP$^{box}_{50}$) as the default evaluation metric. Following the protocol in  \cite{fgfa}, we use motion IoU ($mIoU$) to split the dataset into three groups: slow ($mIoU < 0.7$), medium ($0.7 < mIoU < 0.9$) and fast ($mIoU > 0.9$), and report AP$^{box}_{50}$ on three subsets, respectively.
As shown in \tab{sota}, our reimplemented SELSA with R-101 \cite{resnet} achieves 81.5\% of AP$^{box}_{50}$ at 14.5 FPS. Then, we replace the R-101 backbone with our TDViT-T, and the performance is improved by 2.4\% to 83.9\% of AP$^{box}_{50}$ at 16.2 FPS. We further introduce the advanced variant TDViT$^+$, and our method achieves 84.5\% of $AP^{box}_{50}$ at 15.7 FPS.
Compared to other SOTA methods, our method achieves a better speed-accuracy trade-off. For example, MAMBA \cite{mamba} and TFBlender+MEGA \cite{tfblender} achieve 84.6\% and 83.8\% of AP$^{box}_{50}$, respectively, but they run at relatively slow speeds, 9.1 and 4.9 FPS.

\begin{table}[t]
\centering
\caption{Comparisons with other backbones for video instance segmentation. }
\label{tab:backbones_vis}
\begin{tabular}{lccccccccc} 
\toprule
Backbone & Arch.          & AP$^{mask}$ & AP$^{mask}_{50}$ & AP$^{mask}_{75}$ & AP$^{mask}_S$ & AP$^{mask}_M$ & AP$^{mask}_L$ & \#Param & FPS   \\
\midrule
R-50   \cite{resnet}    & \multirow{2}{*}{CNN}  & 30.3     & 50.0        & 32.1        & 11.8       & 30.5       & 40.6       & 58.1M & 18.0  \\
R-101  \cite{resnet}     &                       & 32.9     & 52.3        & 35.6        & 9.8        & 33.0       & 44.1       & 77.1M  & 17.2  \\
\midrule
Swin-T  \cite{swin}  & \multirow{2}{*}{ViT} & 33.7     & 57.1        & 35.1        & 12.8       & 33.5       & 44.2       & 61.5M  & 18.6  \\
Swin-S  \cite{swin}  &                       & 36.8     & 60.9        & 40.1        & 16.0       & 37.3       & 48.8       & 82.8M  & 14.3  \\
\midrule
TDViT-T  & \multirow{4}{*}{ViT} & 35.4     & 59.6        & 38.6        & 13.6       & 36.9       & 46.4       & 61.5M  & 18.6  \\
TDViT-T$^+$ &                       & 36.1     & 60.2        & 39.9        & 14.0       & 37.5       & 48.1       & 65.0M  & 18.0  \\
TDViT-S  &                       & 37.7     & 61.9        & 40.5        & 16.9       & 37.9       & 49.6       & 82.8M  & 14.3  \\
TDViT-S$^+$ &                       & 38.2     & 62.5        & 41.6        & 17.6       & 38.1       & 51.2       & 86.3M  & 13.9  \\
\bottomrule
\end{tabular}
\end{table}

\subsection{Video Instance Segmentation}
Similar to the experiment designs in video object detection, we first compare TDViT with other backbones using MaskTrack R-CNN \cite{youtubevis} as the base sgementor. Then, we compare the performance with other SOTA video instance segmentation methods.

\subsubsection{Comparisons with other backbones.} 
We compare our TDViT-T with other backbones with similar model sizes, i.e., number of parameters (\#Param) in \tab{backbones_vis}. The pre-trained Mask R-CNN \cite{maskrcnn} model on the COCO \cite{coco} dataset is used to initialise the network.
We first report results using R-50 and R-101 backbones. Then, we replace the backbone with a stronger Swin model \cite{swin}. Compared with using R-50/101, the performance of using Swin-T/S is improved by 3.4/3.9\%  to 33.7/36.8\% of AP$^{mask}$, respectively.
Then, we replace the backbone with TDViT. Compared to Swin-T/S, the performance of TDViT-T/S is improved by 1.7/0.9\%  to 35.4/37.7\% of AP$^{mask}$, respectively.
By introducing the advanced variants, TDViT-T$^+$/S$^+$, the performance is further improved to 36.1/38.2\% of AP$^{mask}$.
We also test the run-time speed of these backbones. As shown in \tab{backbones_vis}, our method can achieve better speed-accuracy trade-offs.

\begin{table}[t]
\centering
\caption{Comparisons of our approach with SOTA methods for video instance segmentation.}
\label{tab:sota_vis}
\begin{tabular}{llccccc} 
\toprule
Methods              & Backbone & AP$^{mask}$   & AP$^{mask}_{50}$ & AP$^{mask}_{75}$ & FPS  & Hardware    \\ 
\midrule
Stem-Seg  \cite{stem-seg}             & R-50      & 30.6 & 50.7 & 33.5 & 12.1 & Titan RTX   \\
SipMask  \cite{sipmask}              & R-50      & 33.7 & 54.1 & 35.8 & 28.0 & Titan RTX   \\
SG-Net  \cite{sgnet}              & R-50      & 34.8 & 56.1 & 36.8 & 22.9 & Titan RTX   \\
TFBlender           \cite{tfblender}  & R-50      & 35.7 & 57.1 & 37.6 & 21.3 & Titan RTX   \\
CrossVis  \cite{crossvis}            & R-50      &36.3 &56.8 &38.9 &- &-\\
VisTR    \cite{vistr}             & R-50      & 36.2 & 59.8 & 36.9 & 30.0 & Tesla V100  \\ 
\hline
MaskTrack R-CNN*  \cite{youtubevis}     & R-50      & 30.3 & 50.0 & 32.1 & 18.0 & Tesla V100  \\
MaskTrack R-CNN+Ours & TDViT-T  & 35.4 & 59.6 & 38.6 & 18.6 & Tesla V100  \\
MaskTrack R-CNN+Ours & TDViT-T$^+$ & 36.1 & 60.2 & 39.9 & 18.0 & Tesla V100  \\
\bottomrule
\end{tabular}
\end{table}

\subsubsection{Comparisons with SOTA.} 
We compare our method with SOTA video instance segmentation methods and report detailed results in \tab{sota_vis}. We use MaskTrack R-CNN  \cite{youtubevis} as the base segmentor and use * to denote our reimplemented version. Using R-50 as the backbone, MaskTrack R-CNN * achieves 30.3\% AP$^{mask}$ at 18.0, which is the baseline.
By replacing the backbone with TDViT-T/T$^+$, the performance is improved to 35.4/36.1\% of AP$^{mask}$ and runs at 18.6/18.0 FPS, respectively. The performance and speed are comparable to recent SOTA methods, e.g., VisTR \cite{vistr} and CrossVis \cite{crossvis}.

\subsection{Ablation Study}
\label{subsec:ablation}

\subsubsection{Spatiotemporal attention schemes.}
We conduct experiments on the split and factorised schemes. Detailed orders of space-only blocks and temporal blocks (TDTB) inside the stage 3 are listed in \tab{spatiotemporal}. Other stages contain 2 blocks, and we use one space-only block followed by one TDTB. More details of spatiotemporal schemes on different variants are shown in the supplementary material. The split scheme has 0.5\% higher AP$^{box}_{50}$ than the factorised scheme, and thus we use the split scheme by default. With a increased number of TDTBs, e.g., from 6 to 8, TDViT-T$^+$ achieves better performance with 79.9\% of AP$^{box}_{50}$.

\begin{table}[t]
\caption{Effect of different spatiotemporal attention schemes.}
\label{tab:spatiotemporal}
\centering
\begin{tabular}{llccc} 
\toprule
\multirow{2}{*}{Scheme} & \multirow{2}{*}{Backbone} & \multicolumn{2}{c}{Stage 3} & \multirow{2}{*}{AP$^{box}_{50}$}  \\
                          &                         & Details    & Depth          &                                   \\ 
\midrule
Space-only     & Swin-T           & {[}s]*6    & 6              & 77.2                              \\
\midrule
Factorised scheme                & TDViT-T              & {[}s, t]*3 & 6              & 78.0                              \\
Split scheme               & TDViT-T                  & s*3, t*3   & 6              & \textbf{78.5}                              \\
\midrule
Split scheme              &    TDViT-T$^+$                 & s*3, t*5   & 8              & \textbf{79.9}                     \\
\bottomrule
\end{tabular}

\end{table}

\subsubsection{Temporal dilations.} 
Larger temporal dilations enable TDViT to fuse temporal information over a longer time span. However, larger temporal dilations increase the difficulty of learning the spatiotemporal correspondences. In this part, we conduct experiments to find a good trade-off between the large temporal receptive field and the learning difficulty. 
We gradually increase the temporal dilations from the lower stage to the higher stage.
The specific number of temporal dilations and the corresponding results are listed in \tab{dilation}. Temporal dilations of four stages are denoted as $\{D_t\}$.
Performance improves when we increase the temporal dilations. In our experiments, with temporal dilations $\{D_t\}=\{4,8,16,32\}$, our method achieves the optimal performance 78.5\% of AP$^{box}_{50}$ and we use this setting by default.
The performance improvement is saturated when we further increase the temporal dilations. For example, with $\{D_t\}=\{5,10,20,40\}$, the performance drops by 0.1\% of AP$^{box}_{50}$.

\begin{table}[t]
\caption{Effect of different temporal dilations. The four numbers denote the temporal dilations in four stages, respectively.}
\label{tab:dilation}
\centering

\begin{tabular}{llllll}
\toprule
$\{D_t\}$    & AP$^{box}_{50}$ & AP$^{box}_{75}$ & AP$^{box}_S$ & AP$^{box}_M$ & AP$^{box}_L$ \\ \midrule
(1,2,4,8)    & 77.2            & 51.5            & 8.6          & 23.2         & 52.4         \\
(2,4,8,16)   & 77.9            & 52.0            & 8.6          & 23.4         & 53.0         \\
(3,6,12,24)  & 78.1            & 52.1            & \textbf{8.7}          & 23.8         & 52.2         \\
(4,8,16,32)  & \textbf{78.5}            & \textbf{52.7}            & 8.0          & \textbf{25.6}         & 53.1         \\
(5,10,20,40) & 78.4            & 52.5            & 8.5          & 24.9         & \textbf{53.4}        
\\
\bottomrule
\end{tabular}
\end{table}

\subsubsection{Memory sampling strategies.}

In the TDTB, a good sampling strategy should be able to generate informative features from the memory structure. For example, given the temporal dilation $D_t$ = 16, the sampled features should well represent 16 neighbour frames.
We design different sampling strategies and conduct experiments to find a better solution. Their differences are illustrated in \S \ref{subsect:tdtb} and \figa{sampling}. The detailed results are shown in \tab{sampling}. 
Among these strategies, patch shuffle achieves the best performance 78.8\% of AP$^{box}_{50}$ because the sampled features have good diversities. However, considering the run-time speed, we use the temporal earliest strategy by default, which achieves a comparable accuracy with 78.5\% of AP$^{box}_{50}$ but a much faster speed at 23.9 FPS.

\begin{table}[t]
\caption{Comparisons between different sampling strategies}
\label{tab:sampling}
\centering
\begin{tabular}{lllllll}
\toprule
                  & 
  AP$^{box}_{50}$ &
  AP$^{box}_{75}$ &
  AP$^{box}_S$ &
  AP$^{box}_M$ &
  AP$^{box}_L$ & FPS  \\
  \midrule
Temporal earliest & 78.5 & 52.7 & 8.0  & 25.6    & 53.1    & \textbf{23.9} \\
Temporal NMS      & 78.5 & 52.6 & 8.3  & \textbf{25.7}    & 53.0    & 19.9 \\
Channel shuffle    & 77.7 &51.4 & 8.4  & 25.1    & 52.1    & 20.2 \\
Patch shuffle      & \textbf{78.8} & \textbf{53.0}  & \textbf{8.5}  & \textbf{25.7}  &\textbf{53.3}     & 19.7\\
\bottomrule
\end{tabular}
\end{table}

\subsubsection{Efficient local attention.}
Local attention is important for TDViT to achieve a good speed-accuracy trade-off. 
We conduct experiments on different local attentions.
Firstly, the window based local attention, which is used in Swin  \cite{swin}. The feature maps are split into non-overlapping windows and then attentions are computed within each window locally. Secondly, the correlation based local attention, which is used in PSLA  \cite{psla}, optical flow  \cite{flownet} and tracking methods  \cite{siamrpn++,siamrpn}. Given one query token, local attentions are computed within a small region around the query location. The differences between these two designs are illustrated in \S \ref{subsect:tdtb} and \figa{st_scheme_and_local_att}. We also test with different local region sizes. The detailed settings and results are presented in \tab{local-att}. By default, we use the window based attention with size 7$\times$7 for a good speed-accuracy trade-off.

\begin{table}[t]
\caption{Comparisons of local attentions according to region sizes.}
\label{tab:local-att}
\centering
\begin{tabular}{ccccc}
\toprule
Size & Window & Correlation  & AP$^{box}_{50}$  & FPS  \\
\midrule
7$\times$7  &        & \checkmark     & \textbf{78.8}      &           18.4 \\
7$\times$7  & \checkmark                  &    & 78.5                & 23.9 \\
5$\times$5  & \checkmark                   &    & 77.6                & \textbf{25.2} \\
9$\times$9  & \checkmark                  &    & 78.1                & 22.1 \\
\bottomrule
\end{tabular}
\end{table}


\section{Conclusion}
\label{sec:con}

We present a Temporal Dilated Video Transformer (TDViT) for dense video tasks. The key component in TDViT is the temporal dilated transformer block (TDTB), which can obtain accurate multi-frame spatiotemporal representations with a single-frame computational cost and effectively sample useful information from redundant frames. Moreover, by using hierarchical TDTBs, our approach can capture long-range temporal correlations, further improving accuracy. 
TDViT achieves excellent speed-accuracy trade-offs on the ImageNet VID and YouTube VIS datasets and is compatible with different frameworks. The simplicity and strong performance suggest that TDViT can potentially serve as a general-purpose backbone for various dense video tasks.

%
%
\bibliographystyle{splncs04}
\bibliography{egbib}
\end{document}